\documentclass[10pt,twocolumn,letterpaper]{article}

\usepackage{cvpr}              

\usepackage{graphicx}
\usepackage{amsmath}
\usepackage{amssymb}
\usepackage{booktabs}

\usepackage{color}
\usepackage{adjustbox}
\usepackage{array}
\usepackage{xcolor,colortbl}
\usepackage{balance}

\newcommand{\Amat}{{\bf A}}

\newcommand{\Dmat}{{\bf D}}

\newcommand{\Hmat}[0]{{{\bf H}}}

\newcommand{\Kmat}[0]{{{\bf K}}}

\newcommand{\Mmat}[0]{{{\bf M}}}

\newcommand{\Qmat}[0]{{{\bf Q}}}
\newcommand{\Rmat}[0]{{{\bf R}}}

\newcommand{\Vmat}[0]{{{\bf V}}}
\newcommand{\Wmat}[0]{{{\bf W}}}
\newcommand{\Xmat}{{\bf X}}
\newcommand{\Ymat}[0]{{{\bf Y}}}
\newcommand{\Zmat}{{\bf Z}}

\newcommand{\xv}{\boldsymbol{x}}
\newcommand{\yv}{\boldsymbol{y}}

\newcommand{\zv}{\boldsymbol{z}}

\usepackage[pagebackref,breaklinks,colorlinks]{hyperref}

\usepackage[capitalize]{cleveref}
\crefname{section}{Sec.}{Secs.}
\Crefname{section}{Section}{Sections}
\Crefname{table}{Table}{Tables}
\crefname{table}{Tab.}{Tabs.}

\def\cvprPaperID{5340} 
\def\confName{CVPR}
\def\confYear{2023}

\begin{document}

\title{EfficientSCI: Densely Connected Network with Space-time Factorization for \\Large-scale Video Snapshot Compressive Imaging}

\author{%
	Lishun Wang $^{1,2,*}$, Miao Cao $^{3,4,}$\thanks{Equal Contribution, $\dagger$ Corresponding Author} ~, and Xin Yuan $^{3, \dagger}$\\
		$^{1}$ Chengdu Institute of Computer Application Chinese Academy of Sciences, \\ 
    $^2$ University of Chinese Academy of Sciences, 
    $^3$ Westlake University, $^4$ Zhejiang University
}
\maketitle


\begin{abstract}
  Video snapshot compressive imaging (SCI) uses a two-dimensional detector to capture consecutive video frames during a single exposure time. 
  Following this, an efficient reconstruction algorithm needs to be designed to reconstruct the desired video frames. 
  Although recent deep learning-based state-of-the-art (SOTA) reconstruction algorithms  have achieved good results in most tasks, they still face the following challenges 
  due to excessive model complexity and GPU memory limitations: 
    1) these models need high computational cost, and 
    2) they are usually unable to reconstruct large-scale video frames at high compression ratios.  
    To address these issues, 
    we develop an {\bf{\em  efficient network}} for video SCI 
    by using {\bf {\em dense connections and space-time factorization mechanism}} within a single residual block,
    dubbed {\bf \emph{EfficientSCI}}.  
    The EfficientSCI network can well establish spatial-temporal correlation by using {\bf {\em convolution in the spatial domain and Transformer in the temporal domain}}, respectively. 
    We are the first time to show that 
    an UHD color video with high compression ratio 
    can be reconstructed from a snapshot 2D measurement 
    using a single end-to-end deep learning model with PSNR above 32 dB.
    Extensive results on both simulation and real data show 
    that our method significantly outperforms all previous SOTA algorithms 
    with better real-time performance. 
    The code is at \url{https://github.com/ucaswangls/EfficientSCI.git}.
\end{abstract}

\begin{figure}[!ht]
    \centering 
    \includegraphics[width=1.\linewidth]{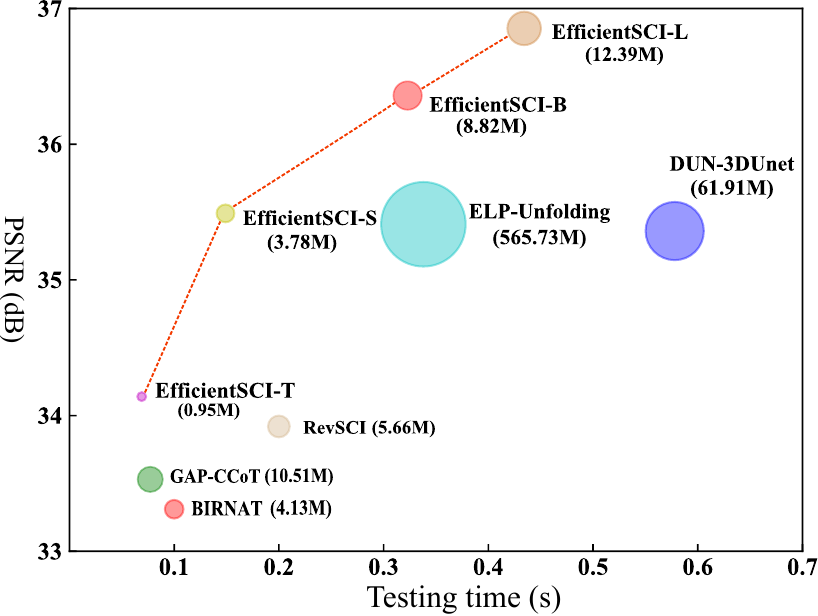}
    \vspace{-5mm}
    \caption{\small{Comparison of reconstruction quality (average PSNR in dB on 6 benchmark grayscale datasets) and testing time of several SOTA deep learning based algorithms. 
      Our proposed EfficientSCI achieves higher reconstruction quality with fewer parameters and shorter testing time.} 
    }
    \vspace{-6mm}
  \label{fig:psnr_time}
\end{figure}
  
\section{Introduction}
\begin{figure}[!ht]
    \centering 
    \includegraphics[width=1.\linewidth]{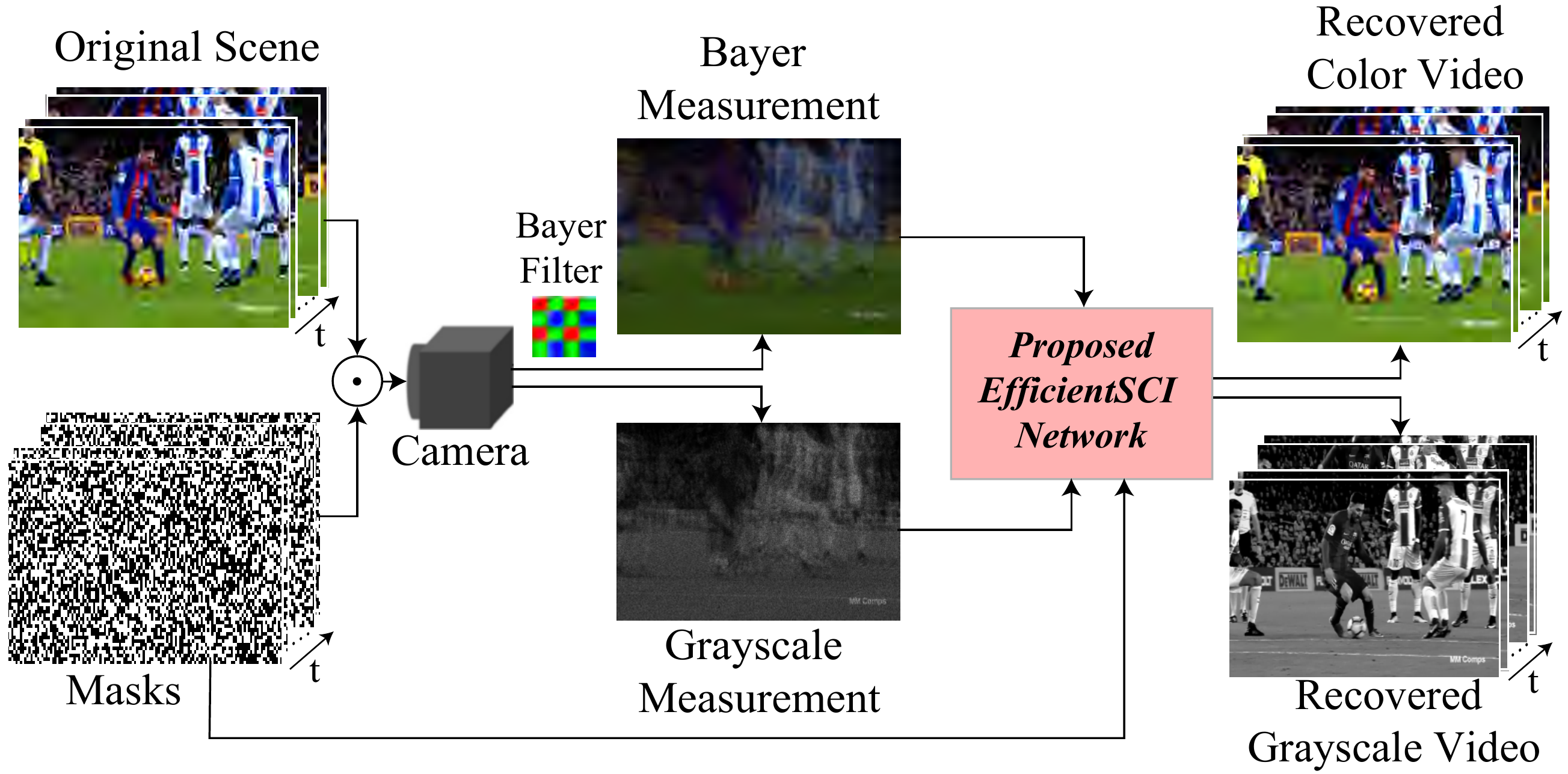}
    \vspace{-6mm}
    \caption{\small{Schematic diagram of grayscale and color video SCI. 
    }
    \vspace{-5mm}
   }
  \label{fig:sci}
\end{figure}

Traditional high-speed camera imaging methods usually suffer from high hardware and storage transmission cost. 
Inspired by compressed sensing (CS) \cite{candes2006robust,donoho2006compressed}, 
video snapshot compressive imaging (SCI) \cite{Yuan2021a} provides an elegant solution. 
As shown in Fig. \ref{fig:sci}, video SCI consists of a hardware encoder and a software decoder. 
In the encoder part, multiple raw video frames are modulated by different masks 
and then integrated by the camera to get a compressed measurement, 
giving low-speed cameras the ability to capture high-speed scenes. 
For the decoding part, the desired high-speed video is retrieved by the reconstruction algorithm using the captured measurement and masks.

So far, many mature SCI imaging systems \cite{Hitomi2011,Reddy2011,Llull2013} have been built, 
but for the decoding part, there are still many challenges. 
In particular, although the model-based methods \cite{Yuan2016,Yang2014,Liu2018} have good flexibility and can reconstruct videos with different resolutions and compression rates, 
they require long reconstruction time and can only achieve poor reconstruction quality. 
In order to improve the reconstruction quality and running speed, 
PnP-FFDNet \cite{Yuan2020c} and PnP-FastDVDnet \cite{yuan2021plug} 
integrate the pre-trained denoising network into an iterative optimization algorithm. 
However, they still need a long reconstruction time on large-scale datasets, \eg, PnP-FastDVDNet takes hours to reconstruct a UHD video from a single measurement. 

By contrast, deep learning based methods \cite{Qiao2020,Meng2020b,Wu2021,wang2022spatial} have better real-time performance and higher reconstruction quality. 
For example, BIRNAT \cite{Cheng2020b} uses bidirectional recurrent neural network 
and generative adversarial method 
to surpass model-based method DeSCI \cite{Liu2018} for the first time.  
MetaSCI \cite{Wang2021e} has made some explorations for the model to adapt to different masks, 
which reduces the model training time. 
DUN-3DUnet \cite{Wu2021} and ELP-Unfolding \cite{Chengshuai} combine iterative optimization ideas with deep learning models to further improve reconstruction quality. 
However, due to the high model complexity and insufficient GPU memory, most existing deep learning algorithms cannot train the models required for reconstructing HD or large-scale videos. 
RevSCI \cite{Cheng2021} uses a reversible mechanism \cite{behrmann2019invertible} to reduce the memory used for model training, 
and can reconstruct HD video with a compression rate up to 24, 
but the model training time increases exponentially. 
In addition, the current reconstruction algorithms generally use convolution to establish spatial-temporal correlation. 
Due to the local connection of convolution, long-term dependencies cannot be well established, 
and the model cannot reconstruct data with high compression rates.

In summary, model-based methods usually require long reconstruction time and can only
achieve poor reconstruction quality. Learning-based methods have high model complexity but cannot be well applied to large-scale color video reconstruction. To address these challenges,  
we develop an {\bf efficient network} for video SCI by using {\em dense connections and space-time factorization mechanism}. 
As shown in Fig. \ref{fig:psnr_time}, our proposed method dramatically outperforms all previous deep learning based reconstruction algorithms 
in terms of reconstruction quality and running speed with fewer parameters. 
Our main contributions can be summarized as follows:
\begin{itemize}
    \item An efficient end-to-end network, dubbed EfficientSCI, is proposed for 
    reconstructing high quality video frames from a snapshot SCI measurement. 
    \item By building hierarchical {\bf d}ense connections 
    within a single {\bf res}idual block, we devise a novel ResDNet block to effectively reduces model computational complexity but enhance the learning ability of the model. 
    \item Based on the {\em space-time factorization} mechanism, a {\bf C}onvolution and Trans{\bf former} hybrid block (CFormer) is built, which can efficiently establish space-time correlation by using convolution in the spatial domain and Transformer in the temporal domain, respectively. 
    \item Experimental results on a large number of simulated and real datasets demonstrate that
    our proposed method achieves state-of-the-art (SOTA) results and better real-time performance. 
\end{itemize}

\section{Related Work}
\begin{figure*}[!ht]
    \centering 
    \includegraphics[width=.95\linewidth]{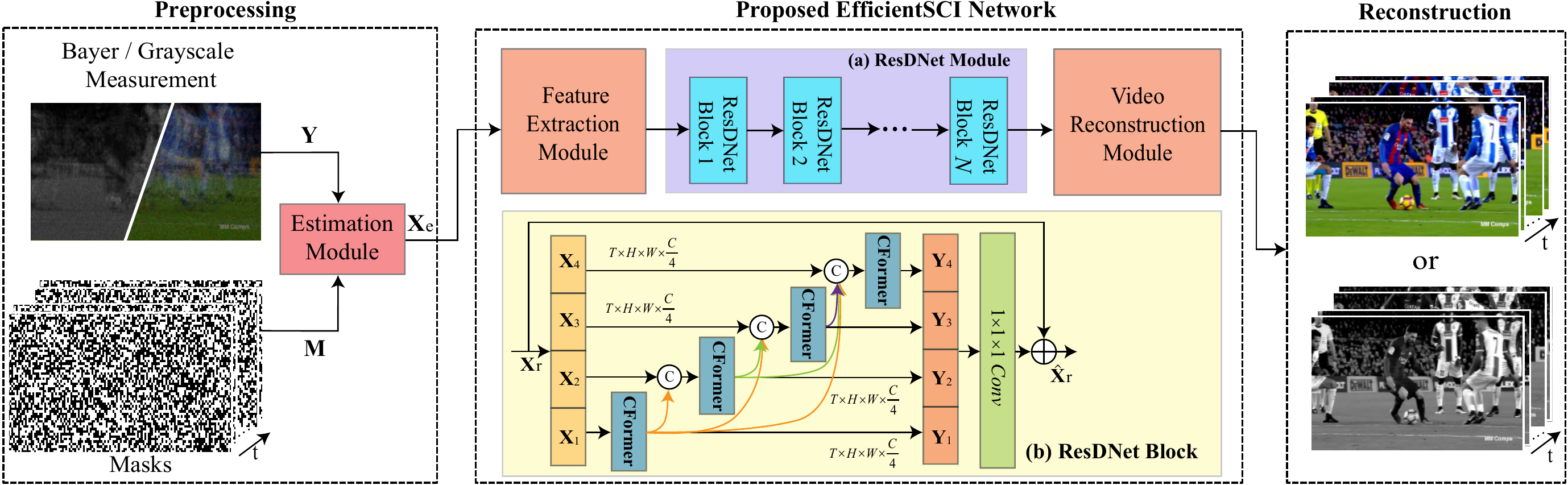}
    \vspace{-3mm}
    \caption{\small{Architecture of the proposed EfficientSCI network and the overall process of color or grayscale video reconstruction. 
    The measurement ($\Ymat$) and masks ($\Mmat$) are pre-processed by the estimation module to obtain an estimated $\Xmat_e$, 
    and then feed $\Xmat_e$ into EfficientSCI network to get the desired reconstruction result. 
    (a) ResDNet module with $N$ residual style units. 
    (b) ResDNet block. Inside a residual block, the input features are divided into $S$ parts by the channel split. Each part uses CFormer to efficiently extract spatial-temporal correlation, and employs dense connections to further improve model capacity.
    For convenience, only the case of $S=4$ is shown here.}
    }
    \vspace{-5mm}
  \label{fig:network}
\end{figure*}

\noindent{\bf CNN and Variants: }
In the past ten years, models with convolutional neural networks (CNN) as the backbone \cite{Krizhevsky2012,He2016,Huang2017}
have achieved excellent results on multiple computer vision tasks \cite{Krizhevsky2012,Redmon2016,Long2015}. 
Among them, ResNeXt \cite{xie2017aggregated} and Res2Net \cite{gao2019res2net} effectively increases model capacity without increasing model complexity by using grouped convolutions inside residual blocks. 
DenseNet \cite{Huang2017} and CSPNet \cite{wang2020cspnet} achieve feature reuse by taking all previous feature maps as input.
In video-related tasks, 
3D convolution can establish good spatial-temporal correlation 
and has been widely used in action recognition \cite{ji20123d}, video super-resolution \cite{luo2020video}, video inpainting \cite{chang2019free} and so on. 
In previous video SCI reconstruction, RevSCI and DUN-3DUnet greatly improve the reconstruction quality of benchmark grayscale datasets by integrating 3D convolution into the network.
However, in complex high-speed scenarios (\eg, \texttt{crash}), since they cannot effectively establish long-term temporal dependencies, 
the reconstruction quality is still lower than 30 dB.
In addition, the excessive use of 3D convolution increases the amount of model parameters, 
which is not conducive to the application of large-scale and high compression ratio data. 

\noindent{\bf Vision Transformers: } 
Most recently, Vision Transformer (ViT) \cite{Dosovitskiy2020} and 
its variants \cite{wang2021pyramid,han2022survey,zamir2022restormer,cai2022mask} have achieved competitive results in computer vision.
However, the high computational complexity limits its application in video-related tasks. 
TimeSformer \cite{bertasius2021space} performs self-attention calculations in time and space respectively, 
which reduces model complexity and improves model accuracy, 
but the computational complexity still increases quadratically with the image size. 
The Video Swin Transformer \cite{ze2021video} limits self-attention calculations to local windows 
but cannot effectively establish long-term temporal dependencies. 
In addition, a large number of experimental results show that Transformer has higher memory consumption than CNN, 
and using Transformer in space is not conducive to large-scale video SCI reconstruction. 
Therefore, through 
{\em space-time factorization mechanism}, 
using Transformer only in time domain can not only effectively utilize its ability 
to establish long-term time series dependencies,  
but also reduce model complexity and memory consumption. 

\section{Mathematical Model of Video SCI}
Fig.~\ref{fig:sci} briefly describes the flow chart of video SCI.
For grayscale video SCI system, the original $B$-frame (grayscale) input video $\{\Xmat_m\}_{m=1}^{B}\in\mathbb{R}^{n_x\times{n_y}}$ is modulated by pre-defined masks $\{\Mmat\}_{m=1}^{B}\in\mathbb{R}^{n_x\times{n_y}}$. 
Then, by compressing across time, 
the camera sensor captures a compressed measurement $\Ymat\in\mathbb{R}^{n_x\times{n_y}}$. 
The whole process can be expressed as:
\begin{equation}
  \Ymat=\textstyle \sum_{m = 1}^{B}{\Xmat_m\odot\Mmat_m+\Zmat},
  \label{eq:Y_mat}
\end{equation}
where $\odot$ denotes the  Hadamard (element-wise) multiplication, and 
$\Zmat\in\mathbb{R}^{n_x\times{n_y}}$ 
denotes the measurement noise. 
Eq. \eqref{eq:Y_mat} can also be represented by a vectorized formulation. 
Firstly, we vectorize $\yv=\operatorname{vec}(\Ymat)\in\mathbb{R}^{n_x{n_y}}$,
$\zv=\operatorname{vec}(\Zmat)\in\mathbb{R}^{n_x{n_y}}$, 
$\xv=\left[\xv_1^{\top},\ldots,\xv^{\top}_{B}\right]^{\top}\in\mathbb{R}^{n_x{n_y{B}}}$,
where $\xv_m=\operatorname{vec}(\Xmat_m)$. 
Then, sensing matrix generated by masks can be defined as: 
\begin{equation}
  \Hmat = \left[\Dmat_1,\ldots,\Dmat_{B}\right]\in\mathbb{R}^{n_x{n_y}\times{n_x{n_y{B}}}}, 
  \label{eq:H}
\end{equation}
where $\Dmat_m=\operatorname{Diag}(\operatorname{vec}(\Mmat)) \in \mathbb{R}^{n_{x} n_{y} \times n_{x} n_{y}}$ is a diagonal matrix and its diagonal elements is filled by $\operatorname{vec}(\Mmat)$. 
Finally, the vectorized expression of Eq. (\ref{eq:Y_mat}) is
\begin{equation}
  \yv = \Hmat{\xv}+\zv. 
  \label{eq:y_vec}
\end{equation}

For color video SCI system, we use the Bayer pattern filter sensor, 
where each pixel captures only red (R), blue (B) or green (G) channel of the raw data in a spatial layout such as `RGGB'. 
Since adjacent pixels are different color components, 
we divide the original measurement $\Ymat$ into four sub-measurements 
$\{\Ymat^{r},\Ymat^{g1},\Ymat^{g2},\Ymat^{b}\}\in\mathbb{R}^{\frac{n_x}{2}\times{\frac{n_y}{2}}}$ 
according to the Bayer filter pattern. 
For color video reconstruction, most of the previous algorithms \cite{Yuan2014,Yuan2020c} reconstruct each sub-measurement independently, 
and then use off-the-shelf demosaic algorithms to get the final  RGB color videos. 
These methods are usually inefficient and have poor reconstruction quality. 
In this paper, we feed the four sub-measurements simultaneously into the reconstruction network to directly obtain the final desired color video.

\section{The Proposed Network}
As shown in Fig. \ref{fig:network}, in the pre-processing stage of EfficientSCI,
inspired by \cite{Cheng2020b,Cheng2021}, we use the estimation module to pre-process  measurement (\Ymat) and  masks (\Mmat) as follows:
\begin{equation}
  \overline{\Ymat}=\textstyle \Ymat\oslash \sum_{m = 1}^{B}{\Mmat_m}, \;
  ~~\Xmat_e = \overline{\Ymat}\odot\Mmat+\overline{\Ymat}, 
  \label{eq:Y_line}
\end{equation}
where $\oslash$ represents Hadamard (element-wise) division, 
$\overline{\Ymat}\in\mathbb{R}^{n_x\times{n_y}}$ is the normalized measurement, which preserves a certain background and motion trajectory information, 
and $\Xmat_e\in\mathbb{R}^{n_x\times{n_y}\times{B}}$ represents the coarse estimate of the desired video. 
We then take $\Xmat_e$ as the input of the EfficientSCI network to get the final reconstruction result. 

EfficientSCI network is mainly composed of three parts: 
$i)$ feature extraction module, $ii)$ ResDNet module and $iii)$ video reconstruction module. 
The feature extraction module is mainly composed of 
three 3D convolutional layers with kernel sizes of $3\times{7}\times{7},  
3\times{3}\times{3}$ and $3\times{3}\times{3}$ respectively. 
Among them, each 3D convolution is followed by a LeakyReLU activation function \cite{maas2013rectifier}, 
and the spatial stride step size of the final 3D convolution is 2. 
The spatial resolution of the final output feature map is reduced to half of the input. 
The feature extraction module effectively maps the input image space to the high-dimensional feature space. 
The ResDNet module is composed of $N$ ResDNet block (described in Sec. \ref{res_b}), which can efficiently explore spatial-temporal correlation.
The video reconstruction module is composed of pixelshuffle \cite{Shi2016} (mainly restore spatial resolution to input network input size) and 
three 3D convolution layers (kernel sizes are 
$3\times{3}\times{3},1\times{1}\times{1}$ and $3\times{3}\times{3}$ respectively), 
and conducts video reconstruction on the features output by the ResDNet blocks.

\subsection{ResDNet Block}
\label{res_b}
Dense connection is an effective way to increase model capacity. Unlike DenseNet, which spans multiple layers, we build a more efficient dense connection within a single residual block. As shown in Fig. \ref{fig:network}(b), the input features of the ResDNet block are first divided into $S$ parts along the feature channel dimension. Then, for each part $i=1,\cdots,S$, we use CFormer (described in Section ~\ref{sec:cformer}) to efficiently establish the spatial-temporal correlation. 
Specifically, for the input of the $i^{th}$ CFormer, we concatenate all the CFormer output features before the $i^{th}$ part with the input features of the $i^{th}$ part and then use a $1\times{1}\times{1}$ convolution to reduce the dimension of the feature channel, which can further reduce the computational complexity. 
Next, we concatenate all CFormer output features along the feature channel dimension and use a $1\times{1}\times{1}$ convolution to better fuse each part of the information. 
Given an input $\Xmat_{r}\in\mathbb{R}^{T\times{H}\times{W}\times{C}}$, ResDNet block can be expressed as:
\begin{align}
\label{Eq:ResDNet_block}
  &\Xmat_{1}, \cdots, \Xmat_{S}= {\rm Split}(\Xmat_{r}),\nonumber\\
  &{\Ymat}_{1} = {\rm CFormer_1}(\Xmat_{1}),\nonumber\\ 
  &{\Ymat}_{2} = {\rm CFormer_2}({\rm Conv_1}({\rm Concat}([{\Ymat}_{1},\Xmat_{2}]))),\nonumber\\
  &\qquad\qquad    \vdots\nonumber\\
  &{\Ymat}_{S} = {\rm CFormer_S}({\rm Conv_1}({\rm Concat}([\Ymat_{1},\cdots,{\Ymat}_{S-1},\Xmat_{S}]))),\nonumber\\
  &\hat{\Ymat}_{r} = {\rm Concat}([{\Ymat}_{1},\cdots,{\Ymat}_{S}]),\nonumber\\
  &\hat{\Xmat}_{r} = {\rm Conv_1(\hat{\Ymat}_r)}+\Xmat_r, 
\end{align}
where `Split' represents division along the channel, 
`${\rm Conv_1}$' represents a $1\times{1}\times{1}$ convolution operation, 
`Concat' represents concatenate along the channel and 
$\hat{\Xmat}_{r}\in\mathbb{R}^{T\times{H}\times{W}\times{C}}$ represents the output of the ResDNet block. 
This design has two advantages: $i)$ the features of different levels are aggregated at a
more granular level, which improves the representation ability of the model; 
$ii)$ the model complexity is reduced (shown in Table  \ref{Tab:ablation_resdnet}).

\begin{figure}[!] 
    \centering 
    \includegraphics[width=1.\linewidth]{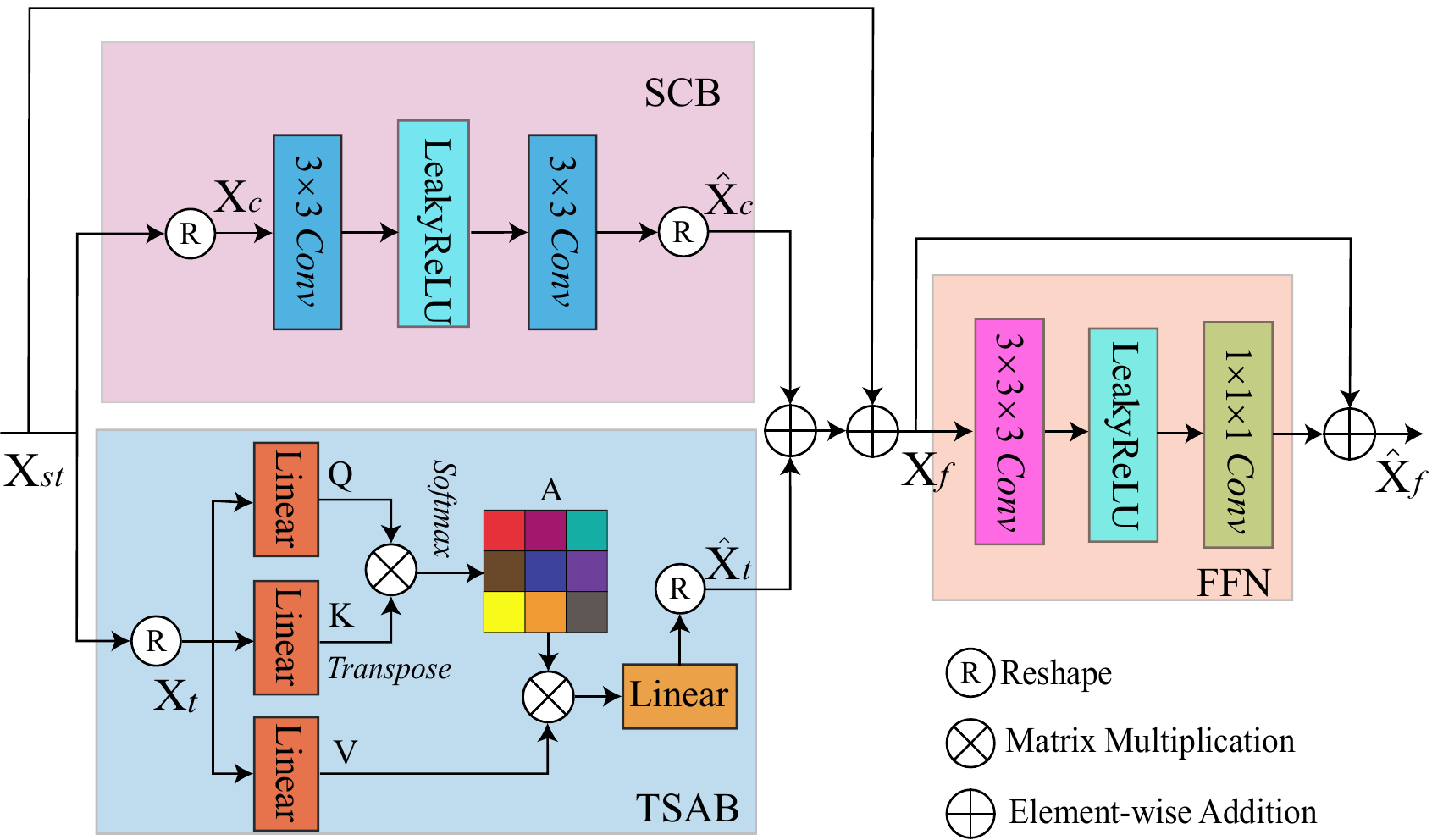}
    \vspace{-4mm}
    \caption{\small{The CFormer block is composed of Spatial Convolution Branch (SCB), Temporal Self-Attention Branch (TSAB) and Feed Forward Network (FFN). 
    For ease of presentation, only the head $N=1$ scenario is described in the TSAB. 
    }
    }
        \vspace{-5mm}
  \label{fig:convoformer}
\end{figure}

\subsection{CFormer Block}
\label{sec:cformer}
As shown in Fig.~\ref{fig:convoformer}, the CFormer block includes three parts: Spatial Convolution Branch (SCB), 
Temporal Self-Attention Branch (TSAB) and Feed Forward Network (FFN). 
Based on the space-time factorization mechanism, SCB is used to extract spatial local information, 
TSAB is used to calculate temporal attention of the feature points at the same spatial position in each frame. 
After that, FFN is used to further integrate spatial-temporal information. 

It is worth noting that in order to make the model flexible to different compression ratios, 
we introduce zero padding position encoding \cite{islam2020much} into CFormer block, 
instead of the absolute position encoding \cite{Dosovitskiy2020} or relative position encoding \cite{Liu2021}. 
Specifically, we modified the first linear transformation layer in the traditional FFN to a $3\times{3}\times{3}$ convolution 
with padding size of 1. 

\noindent{\bf Spatial Convolution Branch:}
2D convolution can well exploit spatial local correlation and reconstructs more detailed information,
and it also enjoys efficient memory consumption and higher operating efficiency, 
which is suitable for large-scale video reconstruction. 
Therefore, We only use two $3\times{3}$ 2D convolutions to reconstruct spatial local details in SCB as shown in Fig. \ref{fig:convoformer}. 

\noindent{\bf Temporal Self-attention Branch: }
The local receptive field of convolution makes it difficult to establish long-term dependencies. 
The global perception ability of Transformer can mitigate this issue. 
However, the time and memory complexity of traditional Transformers increase quadratically with the image size. 
To alleviate this problem, {following \cite{bertasius2021space, wang2022spatial}}, we propose TSAB (shown in Fig. \ref{fig:convoformer}), 
which restricts the self-attention computation to the temporal domain, 
and its complexity only increase linearly with the image/video size. 

In particular, we first reshape the input 
$\Xmat_{st}\in\mathbb{R}^{T\times{H}\times{W}\times{\frac{C}{S}}}$ to 
$\Xmat_t \in\mathbb{R}^{HW\times{T}\times{\frac{C}{S}}}$, 
and then obtain
$query\; (\Qmat\in\mathbb{R}^{HW\times{T}\times{\frac{C}{2S}}})$,
$key\;(\Kmat\in\mathbb{R}^{HW\times{T}\times{\frac{C}{2S}}})$ 
and $value\;(\Vmat\in\mathbb{R}^{HW\times{T}\times{\frac{C}{2S}}})$ by linearly mapping $\Xmat_t$: 
\begin{equation}
  \Qmat = \Xmat_t\Wmat^Q, ~~\Kmat = \Xmat_t\Wmat^K,~~ \Vmat = \Xmat_t\Wmat^V, 
\end{equation}
where $\{\Wmat^Q,\Wmat^K,\Wmat^V\}\in\mathbb{R}^{\frac{C}{S}\times{\frac{C}{2S}}}$ 
are projection matrices. 

It is worth noting that 
the output dimension of the projection matrix is reduced to half of the input dimension, 
further decreasing the computational complexity of TSAB. 
Then, we respectively divide $\Qmat$, $\Kmat$, $\Vmat$ into $N$ heads along the feature channel: 
$\Qmat=\{\Qmat_j\}_1^N$, 
$\Kmat=\{\Kmat_j\}_1^N$, 
$\Vmat=\{\Vmat_j\}_1^N\in\mathbb{R}^{HW\times{T}\times{\frac{C}{2SN}}}$. 
For each head $j=1,\cdots,N$, the attention can be calculated as: 
\begin{equation}
  head_j = \Amat_j*\Vmat_j, 
\end{equation}
where 
$\Amat_j = softmax(\Qmat_{j}\Kmat_{j}^{T}/\sqrt{d})\in\mathbb{R}^{HW\times{T}\times{T}}$ represents an attention map, 
$\Kmat_{j}^{T}$ represents the transposed matrix of $\Kmat_{j}$ and 
$d=\frac{C}{2SN}$ is a scaling parameter. 
Then, we concatenate the outputs of $N$ heads along the feature channel dimension and perform a linear mapping to obtain the final output $\hat{\Xmat}_t\in\mathbb{R}^{T\times{H}\times{W}\times{\frac{C}{S}}}$
of TSAB: 
\begin{eqnarray}
  \hat{\Xmat}_t=\Rmat(\Wmat^P({\rm Concat}[head_1,\cdots,head_N])),
\end{eqnarray}
where $\Wmat^P\in\mathbb{R}^{\frac{C}{2S}\times\frac{C}{S}}$ represents projection matrices, 
and $\Rmat$ is the  reshape operator. 

\begin{table}[!ht]
  \setlength\tabcolsep{3pt}
  \renewcommand{\arraystretch}{1.0}
  \caption{\small{Computational complexity of 
    several SOTA methods.}}
  \vspace{-2mm}
  \centering
  \resizebox{.35\textwidth}{!}
  {
  \centering
  \begin{tabular}{c|c}
  \toprule
  Method & Computational Complexity
  \\
  \midrule
  SCB3D & $\frac{1}{2}HWTK^3C^2$
  \\
  G\mbox{-}MSA & $HWTC^2+(HWT)^{2}C$ 
  \\
  TS\mbox{-}MSA &$2HWTC^2+T(HW)^{2}C+HWT^2C$
  \\
  \midrule
  SCB & $\frac{1}{2}HWTK^2C^2$ 
  \\
  TSAB  & $\frac{1}{2}HWTC^2+\frac{1}{2}HWT^{2}C$
  \\
  \bottomrule
  \end{tabular}
  }
  \label{Tab:complex}
  \vspace{-2mm}
\end{table}
We further analyze the computational complexity of SCB and TSAB, and 
compare them with 3D convolution and 
several classic Multi-head Self-Attention (MSA) mechanisms.
The results are shown in Table  \ref{Tab:complex}, where `SCB3D' represents the replacement of 2D convolution in SCB with 3D convolution {and $K$ represents the kernel size} , 
`G-MSA' represents the original global MSA \cite{Dosovitskiy2020}, 
and `TS-MSA' represents the MSA in TimeSformer \cite{bertasius2021space}.  
It can be observed that the computational complexity of our proposed SCB and TSAB 
grows linearly with the spatial size $HW$, 
the computational cost is much less than `TS-MSA' and `G-MSA' (grow quadratically with $HW$). 
Compared with 3D convolution, since $T$ is generally smaller than $C$, 
`SCB' and `TSAB' still need less computational cost.

\noindent{\bf{Feed Forward Network}: }
The feed forward network of traditional Transformer usually uses two linear layers and a nonlinear activation function to learn more abstract feature representations. 
However, in the whole FFN, there is no interaction between the feature points. 
In order to better integrate the spatial-temporal information and position coding information, 
we replace the first linear transformation layer in the traditional FFN with a $3\times{3}\times{3}$ convolution. 

Given $\Xmat_f\in\mathbb{R}^{T\times{H}\times{W}\times{\frac{C}{2}}}$, FFN can be expressed as:
\begin{equation}
  \hat{\Xmat}_f=\Xmat_f+\Wmat_1(\phi(\Wmat_2(\Xmat_f))),
\end{equation}
where $\Wmat_1, \Wmat_2$ represent $1\times{1}\times{1}$ convolution and $3\times{3}\times{3}$ convolution operations respectively, 
$\phi$ denotes the LeakyReLU non-linearity activation, 
and $\hat{\Xmat}_f \in\mathbb{R}^{T\times{H}\times{W}\times{\frac{C}{2}}}$ is the output of the FFN. 

It should be noted that in the whole CFormer block, we do not use any regularization layers, 
such as Layer Normalization \cite{ba2016layer} and Batch Normalization \cite{ioffe2015batch}. 
The experimental results show that 
{\em removing the regularization layer will 
not reduce the quality of model reconstruction and can further improve the efficiency of the model.}

\begin{table}[!ht]
    \setlength\tabcolsep{3pt}
  \renewcommand{\arraystretch}{1.0}
  \caption{\small{Reconstruction quality and test time (s) 
    using EfficientSCI with different number of channels and blocks.}}
    \vspace{-3mm}
  \centering
  \resizebox{.4\textwidth}{!}
  {
  \centering
  \begin{tabular}{c|ccccc}
    \toprule
    Model & Channel & Block & PSNR & SSIM &Test time(s)
    \\
    \midrule
    EfficientSCI-T & 64 & 8 & 34.22 & 0.961 & 0.07
    \\
    EfficientSCI-S & 128 & 8 & 35.51 & 0.970 & 0.15
    \\
    EfficientSCI-B & 256 & 8 & 36.48 & 0.975 & 0.31
    \\
    EfficientSCI-L & 256 & 12 & 36.92 & 0.977 & 0.45
    \\
    \bottomrule
  \end{tabular}
  \vspace{-3mm}
  } 
  \label{Tab:chan_block}
\end{table}

\begin{table}[!ht]
  \vspace{-2mm}
  \setlength\tabcolsep{3pt}
  \renewcommand{\arraystretch}{1.0}
  \caption{\small{Computational complexity and reconstruction quality of several SOTA algorithms on 6 grayscale benchmark datasets.}}
  \vspace{-3mm}
  \centering
  \resizebox{.35\textwidth}{!}
  {
  \centering
  \begin{tabular}{c|cccc}
  \toprule
  Method & Params (M) &FLOPs (G) &PSNR &SSIM
  \\
  \midrule
  BIRNAT &4.13 &390.56&33.31&0.951
  \\
  RevSCI  &5.66 &766.95  &33.92 &0.956
  \\
  DUN-3DUnet &61.91 &3975.83 &35.26 &0.968
  \\
  ELP-Unfolding &565.73 &4634.94 &35.41 &0.969
  \\
  \midrule
  EfficientSCI-T &0.95 &142.18 &34.22 &0.961
  \\
  EfficientSCI-S &3.78 &563.87 &35.51 &0.970
  \\
  EfficientSCI-B &8.82 &1426.38 &36.48 &0.975
  \\
  EfficientSCI-L &12.39 &1893.72 &36.92 &0.977
  \\
  \bottomrule
  \end{tabular}
  }
  \label{Tab:para_floats}
  \vspace{-4mm}
\end{table}

\noindent{\bf Network Variants:}
To balance speed and performance of the proposed network,
we introduce four different versions of EfficientSCI network, dubbed as EfficientSCI-T,
EfficientSCI-S, EfficientSCI-B and EfficientSCI-L standing for \texttt{Tiny, Small, Base and
Large} networks, respectively.
The network hyper-parameters are shown in Table \ref{Tab:chan_block}, in which
we mainly changed the the number of ResDNet blocks and the number of channels. 
As shown in Table \ref{Tab:para_floats}, we also compare model parameters and computational complexity (FLOPs) with several advanced methods. The complexity of our proposed EfficientSCI-T is smaller than that of BIRNAT and RevSCI, 
and EfficientSCI-L is smaller than that of DUN-3DUnet and ELP-Unfolding. 
\begin{table*}[!htbp]
  \renewcommand{\arraystretch}{1.0}
  \caption{\small{The average PSNR in  dB (left entry) and SSIM (right entry) and running time per measurement of different algorithms on 6 benchmark grayscale datasets. 
  The best results are shown in bold and the second-best results are underlined.}}
  \vspace{-3mm}
  \centering
  \resizebox{.99\textwidth}{!}
  {
  \centering
  \begin{tabular}{c|cccccccc}
  \toprule
  Method 
  & Kobe 
  & Traffic 
  & Runner 
  & Drop 
  & Crash 
  & Aerial 
  & Average 
  & Test time(s) 
  \\
  \midrule
  GAP-TV
  & 26.46, 0.885
  & 20.89, 0.715
  & 28.52, 0.909
  & 34.63, 0.970 
  & 24.82, 0.838 
  & 25.05, 0.828 
  & 26.73, 0.858
  & 4.2 (CPU) 
  \\
  PnP-FFDNet
  & 30.50, 0.926  
  & 24.18, 0.828
  & 32.15, 0.933
  & 40.70, 0.989
  & 25.42, 0.849
  & 25.27, 0.829
  & 29.70, 0.892
  & 3.0 (GPU)
  \\
  PnP-FastDVDnet
  & 32.73, 0.947
  & 27.95, 0.932 
  & 36.29, 0.962 
  & 41.82, 0.989 
  & 27.32, 0.925 
  & 27.98, 0.897
  & 32.35, 0.942
  & 6.0 (GPU)
  \\
  DeSCI
  & 33.25, 0.952
  & 28.71, 0.925
  & 38.48, 0.969
  & 43.10, 0.993
  & 27.04, 0.909
  & 25.33, 0.860
  & 32.65, 0.935
  & 6180 (CPU)
  \\
  BIRNAT
  & 32.71, 0.950
  & 29.33, 0.942 
  & 38.70, 0.976
  & 42.28, 0.992
  & 27.84, 0.927
  & 28.99, 0.917
  & 33.31, 0.951
  & 0.10 (GPU)
  \\
  RevSCI
  &33.72, 0.957
  &30.02, 0.949
  &39.40, 0.977
  & 42.93, 0.992
  & 28.12, 0.937
  & 29.35, 0.924
  & 33.92, 0.956
  & 0.19 (GPU)
  \\
  GAP-CCoT
  & 32.58, 0.949
  & 29.03, 0.938
  & 39.12, 0.980
  & 42.54, 0.992 
  & 28.52, 0.941
  & 29.40, 0.923
  & 33.53, 0.958
  & \underline {0.08 (GPU)}
  \\
  DUN-3DUnet
  & 35.00, 0.969
  & 31.76, 0.966
  & 40.03, 0.980
  & 44.96, 0.995
  & 29.33, 0.956
  & 30.46, 0.943
  & 35.26, 0.968
  & 0.58 (GPU)
  \\
  ELP-Unfolding
  & 34.41, 0.966
  & 31.58, 0.962
  & 41,16, 0.986
  & 44.99, 0.995
  & 29.65, 0.959 
  & 30.68, 0.944
  & 35.41, 0.969
  & 0.34 (GPU)
  \\
  \midrule
  EfficientSCI-T
  & 33.45, 0.960
  & 29.20, 0.942
  & 39.51, 0.981
  & 43.56, 0.993
  & 29.27, 0.954
  & 30.32, 0.937
  & 34.22, 0.961
  & {\bf 0.07 (GPU)}
  \\
  EfficientSCI-S
  & 34.79, 0.968
  & 31.21, 0.961
  & 41.34, 0.986
  & 44.61, 0.994
  & 30.34, 0.965
  & 30.78, 0.945
  & 35.51, 0.970
  & 0.15 (GPU)
  \\
  EfficientSCI-B
  & \underline{35.76}, \underline{0.974}
  & \underline{32.30}, \underline{0.968}
  & \underline{43.05},\underline{ 0.988}
  & \underline{45.18},\underline{ 0.995}
  & \underline{31.13},\underline{ 0.971}
  & \underline{31.50}, \underline{0.953}
  & \underline{36.48}, \underline{0.975}
  & 0.31 (GPU)
  \\
  EfficientSCI-L
  & {\bf 36.27}, {\bf 0.976}
  & {\bf 32.83}, {\bf 0.971}
  & {\bf 43.79}, {\bf 0.991}
  & {\bf 45.46}, {\bf 0.995}
  & {\bf 31.52}, {\bf 0.974}
  & {\bf 31.64}, {\bf 0.955}
  & {\bf 36.92}, {\bf 0.977}
  & 0.45 (GPU)
  \\
  \bottomrule
  \end{tabular}
  }
  \vspace{-3mm}
  \label{Tab:sim6}
\end{table*}

\section{Experiment Results}
\subsection{Datasets}
Following BIRNAT \cite{Cheng2020b}, we use \texttt{DAVIS2017} \cite{pont2017} with resolution
$480\times{894}$ (480p) as the model training dataset. 
To verify model performance, we first test the EfficientSCI network on several simulated datasets,  
including six benchmark grayscale datasets 
(\texttt{Kobe, Traffic, Runner, Drop, Crash} and \texttt{Aerial} with a size of $256\times{256}\times{8}$), 
six benchmark mid-scale color datasets
(\texttt{Beauty, Bosphorus, Jockey, Runner, ShakeNDry} and \texttt{Traffic} with a size of $512\times{512}\times{3}\times{8}$), 
and four large-scale datasets 
(\texttt{Messi, Hummingbird, Swinger} and \texttt{Football} with different sizes and compression ratios). 
Then we test our model on some real data (including \texttt{Domino, Water Balloon}) captured by a real SCI system \cite{Qiao2020}.

\begin{figure}[!ht]
    \centering 
    \includegraphics[width=1.\linewidth]{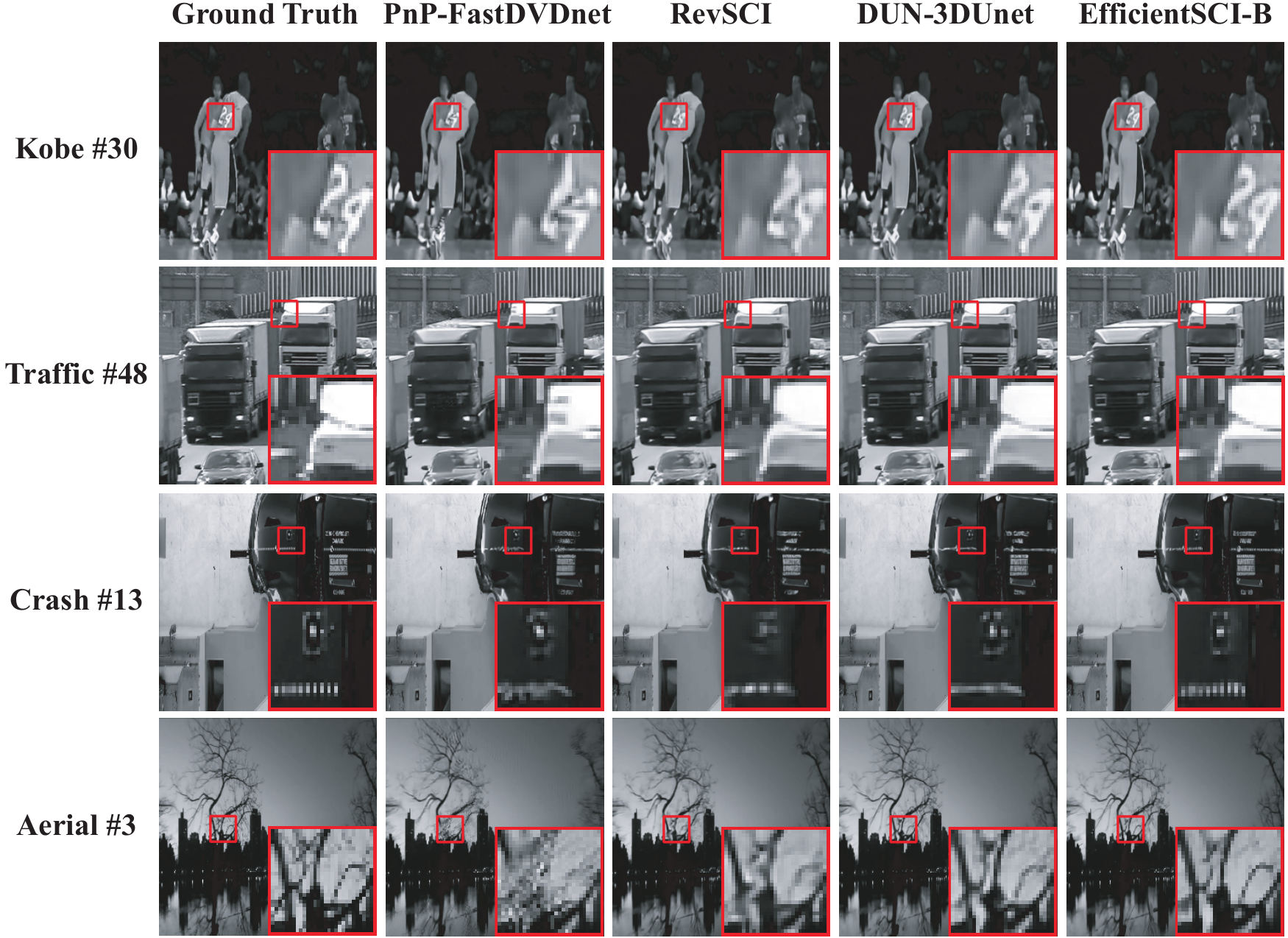}
    \vspace{-6mm}
    \caption{\small{Selected reconstruction frames of simulated grayscale data. Zoom in for better view.}  
    }
\vspace{-6mm}
  \label{fig:gray_sim}
\end{figure}

\subsection{Implementation Details}
We use PyTorch framework with 4 NVIDIA RTX 3090 GPUs for training with random cropping, random scaling, and random flipping for data augmentation,  
and use Adam \cite{Kingma2014} to optimize the model with the initial learning rate 0.0001. 
After iterating for 300 epochs, we adjusted the learning rate to 0.00001 and continued to 
iterate for 40 epochs to get the final model parameters. 
The peak-signal-to-noise-ratio (PSNR) and structured similarity index
metrics (SSIM) \cite{Wang2004} are used as the performance indicators of reconstruction quality. 

\subsection{Results on Simulation Datasets}
\subsubsection{Grayscale Simulation Video}
We compare our method with SOTA model-based methods (GAP-TV \cite{Yuan2016}, PnP-FFDNet \cite{Yuan2020c}, PnP-FastDVDnet \cite{yuan2021plug}, DeSCI \cite{Liu2018})
and deep learning-based methods (BIRNAT \cite{Cheng2020b}, RevSCI \cite{Cheng2021}, GAP-CCoT \cite{wang2022snapshot}, DUN-3DUnet \cite{Wu2021}, ELP-Unfolding \cite{Chengshuai})
on simulated grayscale datasets. 
Table \ref{Tab:sim6} shows the quantitative comparison results,  
it can be observed that our proposed EfficientSCI-L can achieve the highest reconstruction quality 
and has good real-time performance. 
In particular, the PSNR value of our method surpasses the existing best method ELP-Unfolding by 1.46 dB on average.
In addition, our proposed EfficientSCI-T achieves high reconstruction quality while achieving the best real-time performance. 
It is worth noting that, for a fair comparison, 
we uniformly test the running time of all deep learning based methods on the same NVIDIA RTX 3090 GPU.
Fig. \ref{fig:gray_sim} shows the visual reconstruction results of some data. By zooming in  some local areas, 
we can observe that our method can recover sharper edges and more detailed information 
compared to previous SOTA methods. 
The mid-scale color results are shown in the SM due to space limitation and our method outperforms previous SOTA by 2.02 dB in PSNR on the benchmark dataset~\cite{yuan2021plug}.

\begin{table*}[!ht]
  \renewcommand{\arraystretch}{1.0}
  \caption{\small{The average PSNR in dB (left entry) and SSIM (middle entry) and test time (minutes) per measurement (right entry) of different algorithms
  on 4 benchmark large-scale datasets. Best results are in bold.} }
  \centering
  \resizebox{.98\textwidth}{!}
  {
  \centering
  \begin{tabular}{c|ccccc}
  \toprule
  Dataset 
  & Pixel resolution 
  & GAP-TV
  & PnP-FFDNet-color
  & PnP-FastDVDnet-color
  & EfficientSCI-S
  \\
  \midrule
  Messi
  & $1080\times{1920}\times{3}\times{8}$
  & 25.20, 0.874, 0.66
  & 34.28, 0.968, 14.93
  & 34.34, 0.970, 15.94
  & {\bf 34.41, 0.973, 0.09}
  \\
  Hummingbird
  & $1080\times{1920}\times{3}\times{30}$
  & 25.10, 0.750, 20.3
  & 28.79, 0.665, 61.20
  & 31.17, 0.916, 54.00
  & {\bf 35.56, 0.952, 0.39}
  \\
  Swinger
  & $2160\times{3840}\times{3}\times{15}$
  & 22.68, 0.769, 39.2
  & 29.30, 0.934, 138.8
  & 30.57, 0.949, 138.4
  & {\bf 31.05, 0.951, 0.62}
  \\
  Football
  & $1644\times{3840}\times{3}\times{40}$
  & 26.19, 0.858, 83.0
  & 32.70, 0.951, 308.8
  & 32.31, 0.947, 298.1
  & {\bf 34.81, 0.964, 1.52}
  \\
  \bottomrule
  \end{tabular}
  }
  \label{Tab:large_any_cr_psnr}
\end{table*}
\begin{figure*}[!h]
    \centering 
    \includegraphics[width=.98\textwidth]{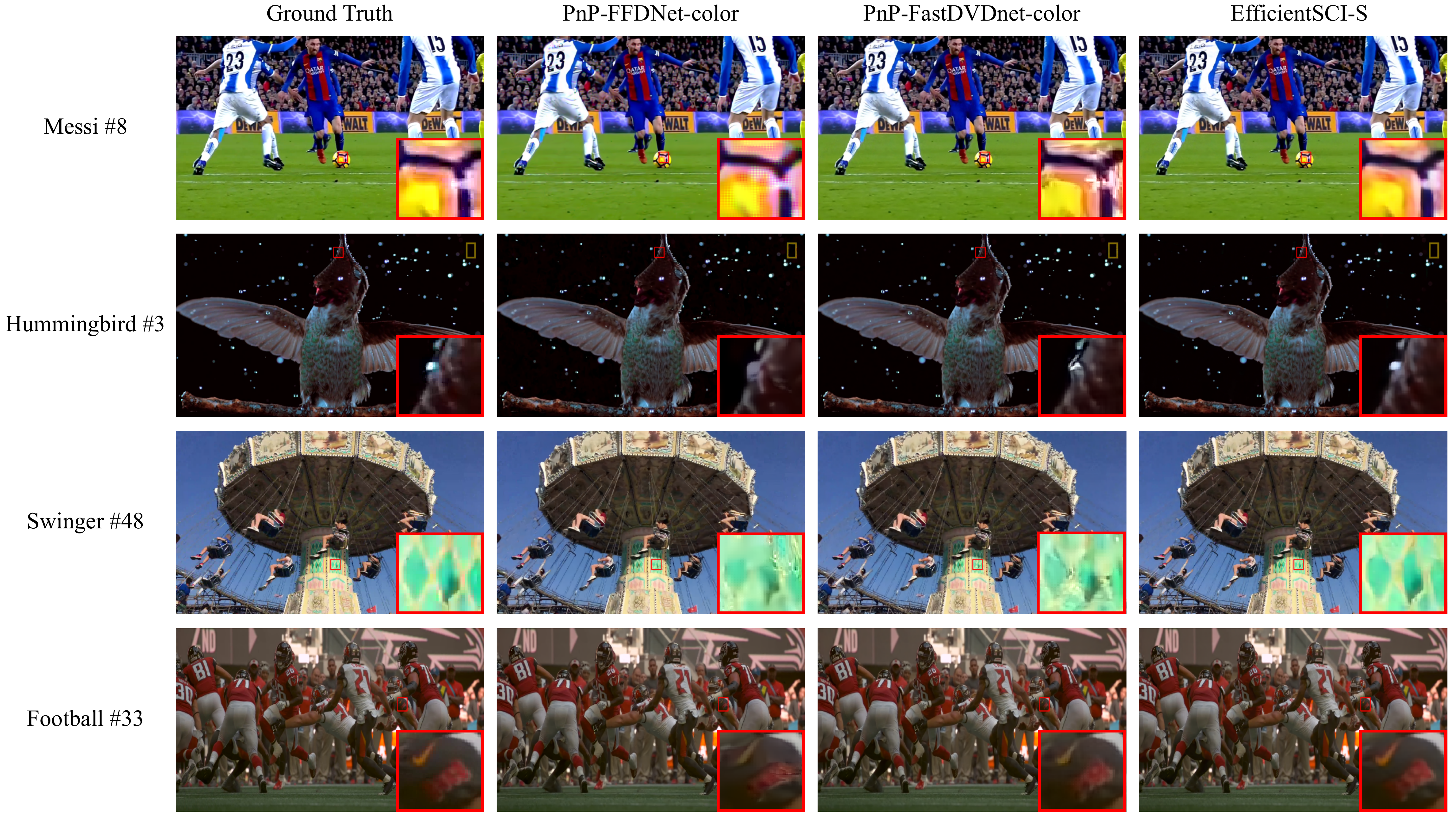}
       \vspace{-2mm}
    \caption{\small{Comparison of reconstruction results of different algorithms 
    on several benchmark large-scale color video simulation datasets.
    }
    }
     \vspace{-2mm}
  \label{fig:large_scale}
\end{figure*}

\begin{figure}[!]
    \centering 
    \includegraphics[width=.45\textwidth]{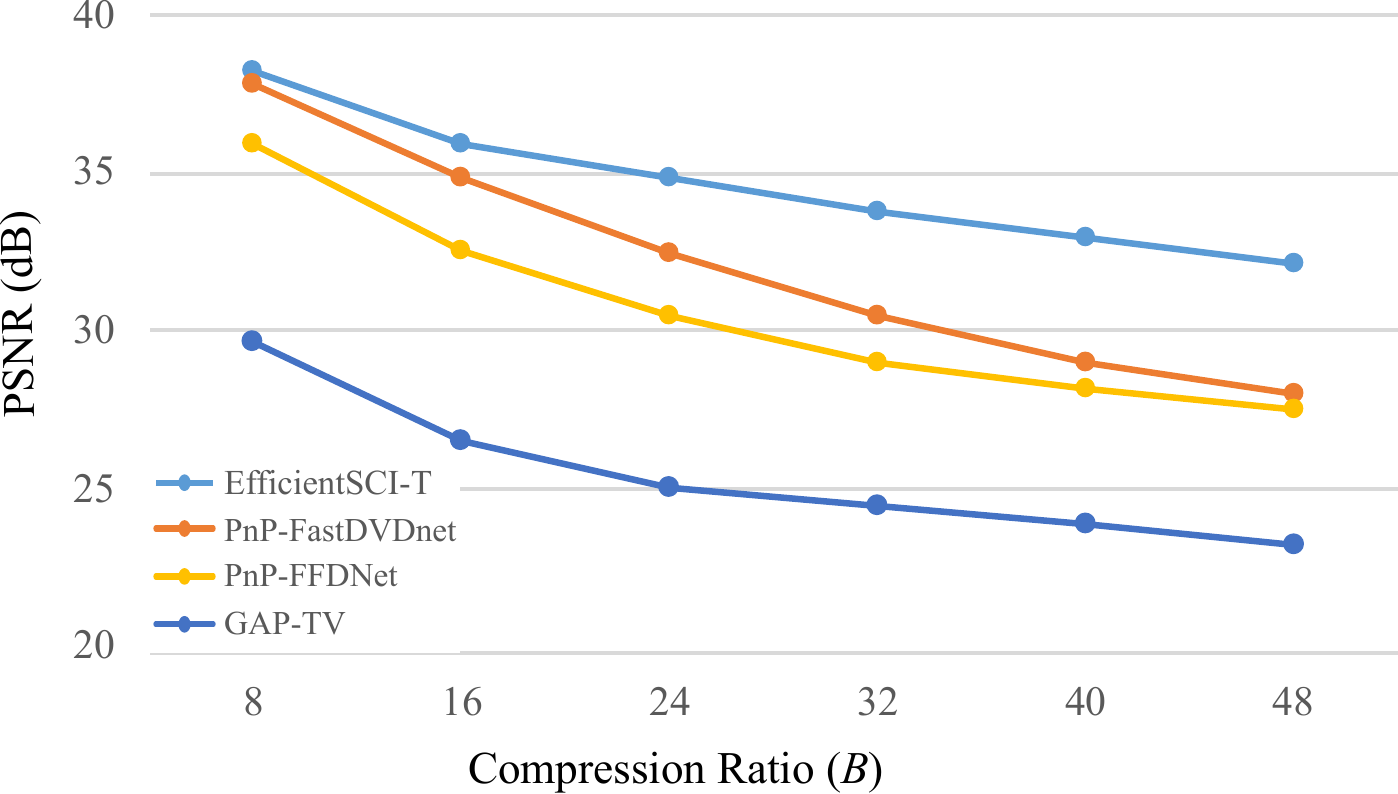}
     \vspace{-3mm}
    \caption{\small{Reconstruction quality
    (PSNR in dB, 
    higher is better) of different reconstruction algorithms, with varying compression rates $B$ from 8 to 48. 
}
    }
    \vspace{-6mm}
  \label{fig:Hummingbird_psnr}
\end{figure}

\subsubsection{Large-scale Color Simulation Video}
Most deep learning based methods, such as BIRNAT \cite{Cheng2020b}, DUN-3DUnet \cite{Wu2021}, cannot be applied to large-scale data reconstruction due to excessive model complexity and GPU memory constraints. 
RevSCI \cite{Cheng2021} uses a reversible mechanism and can reconstruct a 24 frames RGB color video 
with size of $1080\times{1920}\times{3}$, 
but training the model is extremely slow. 
GAP-CCoT \cite{wang2022snapshot} and ELP-Unfolding \cite{Chengshuai} only use 2D convolution for reconstruction and thus cannot handle color video data well. 
Therefore, we only compare with several SOTA model-based methods 
(GAP-TV \cite{Yuan2016}, PnP-FFDNet-color \cite{Yuan2020c}, PnP-FastDVDnet-color \cite{yuan2021plug}) 
on large-scale color data. 
Table \ref{Tab:large_any_cr_psnr} shows the comparisons between our proposed method and several model-based methods on PSNR, SSIM and test time (in minutes).
It can be observed that model-based methods either have long reconstruction time (PnP-FFDNet-color, PnP-FastDVDnet-color) 
or low reconstruction quality (GAP-TV). 
Our proposed EfficientSCI-S can achieve higher reconstruction quality and running speed. 
Especially on UHD color video \texttt{football} ($1644\times{3840}\times{3}\times{40}$), the PSNR value of our method is 2.5 dB higher than PnP-FastDVDnet-color, 
and the reconstruction time is only $0.5\%$ of it. 
Fig. \ref{fig:large_scale} shows some visual reconstruction results. 
By zooming in local areas, we can observe that the reconstruction results of our method are closer to the real value. 
In addition, our proposed model enjoys high flexibility for different compression ratios, that is, the model trained on low compression ratio data can be directly used for high compression ratio video reconstruction task. To verify this,
we test \texttt{hummingbird} data with different compression ratios 
$B=8,16,24,32,40,48$, and the reconstruction results are shown in Fig. \ref{fig:Hummingbird_psnr}. 
We can observe that our method can be applied to video data with different compression ratios, even when the compression ratio $B$ grows to 48, the PSNR value of EfficientSCI-T model can still reach more than 32 dB. Moreover, our proposed approach surpasses other reconstruction algorithms at all compression ratios. 

\subsubsection{Ablation Study}
To verify the performance of the proposed 
ResDNet block and CFormer block on the impact of the reconstruction quality, 
we conduct some ablation experiments. 
The results are shown in Table \ref{Tab:ablation_resdnet} and Table  \ref{Tab:ablation_gf}, 
we not only compare the reconstruction quality of different models, 
but also analyze the model parameters and FLOPs. 
All experiments are conducted on the 6 grayscale benchmark datasets. 

\begin{table}[!ht]
    \vspace{-2mm}
     \caption{\small{Ablation study on the ResDNet block without dense connections (left entry) and with dense connections (right entry).}}
      \vspace{-2mm}
  \centering
  \resizebox{.48\textwidth}{!}
  {
  \begin{tabular}{cccccc}
    \toprule
    GN & Params &FLOPs& PSNR & SSIM
    \\
    \midrule
    1& 27.40, 27.40 & 3860.94, 3860.94 &35.17, 35.17 & 0.967, 0.967
    \\
    2& 14.82, 15.08 & 2211.77, 2246.39 & 35.09, 36.02 &0.966, 0.974
    \\
    4& 8.53, 8.82& 1387.33, 1426.38 &35.02, 36.48& 0.966, 0.975 
    \\
    8& 5.38, 5.65 & 975.39, 1013.45 &34.31, 35.68 & 0.961, 0.971 
    \\
    \bottomrule
  \end{tabular}
  }
    \label{Tab:ablation_resdnet}
\end{table}

\begin{table}[!ht]
    \vspace{-2mm}
  \renewcommand{\arraystretch}{1.0}
  \caption{\small{Ablation study on the CFormer block.}}
  \centering
  \vspace{-2mm}
  \resizebox{.48\textwidth}{!}
  {
  \centering
  \begin{tabular}{cccccccc}
  \toprule
  SCB
  & TSAB
  & Swin
  & S2-3D
  & Params (M)
  & FLOPs (G) 
  & PSNR 
  & SSIM
  \\
  \midrule
  & \checkmark  & \checkmark &  &
  2.88 & 471.03 & 
  34.99 &0.967
  \\
   &  & &\checkmark  & 
    6.93 & 999.31 & 
    34.93 & 0.966
  \\
  \checkmark& \checkmark & & &
      3.78& 563.87 & 
    35.51 &0.970
  \\
  \bottomrule
  \end{tabular}
  }
  \label{Tab:ablation_gf}
    \vspace{-2mm}
\end{table}

\noindent{\bf ResDNet Block: }
We verify the effect of different group numbers (GN) (corresponding to $S$ in Eq. \ref{Eq:ResDNet_block}) and dense connections on the reconstruction quality. As shown in Table \ref{Tab:ablation_resdnet}, the model complexity decreases gradually with the increase of GN, but the reconstruction quality greatly decreases when there are no dense connections in the ResDNet block.
By introducing dense connections in the ResDNet block, the reconstruction quality of our proposed method is greatly improved, and a gain of 1.46 dB can be obtained when GN is 4.

\noindent{\bf CFormer Block: }
In the CFormer block, we first replace SCB with Swin Transformer (Swin) to verify its effectiveness. Then, we replace SCB and TSAB with two stacked 3D convolutions (S2-3D) to verify the effectiveness of TSAB.
As shown in Table \ref{Tab:ablation_gf},  
compared with Swin Transformer, SCB can bring a 0.52 dB gain. Although the number of parameters and FLOPs have increased,  the experimental results show that SCB takes up less memory than the Swin Transformer, 
which is very important for large-scale and high compression ratio data. 
Please refer to more detailed analysis in SM. 
Compared with SCB and TSAB, S2-3D not only increases model parameters and FLOPs by $83\%$ and $77\%$ respectively, but also reduces the PSNR value by 0.58 dB, which verifies the necessity of using space-time factorization and TSAB.

\noindent{\bf Number of Channels and Blocks:} 
Table \ref{Tab:chan_block} shows that the quality of model reconstruction increases with the number of channels and blocks.
However, the amount of parameters and FLOPs also increase (see Table ~\ref{Tab:para_floats}), 
resulting in a degradation in the real-time performance of the model.
\begin{figure}[!ht]
    \centering 
    \includegraphics[width=1.\linewidth]{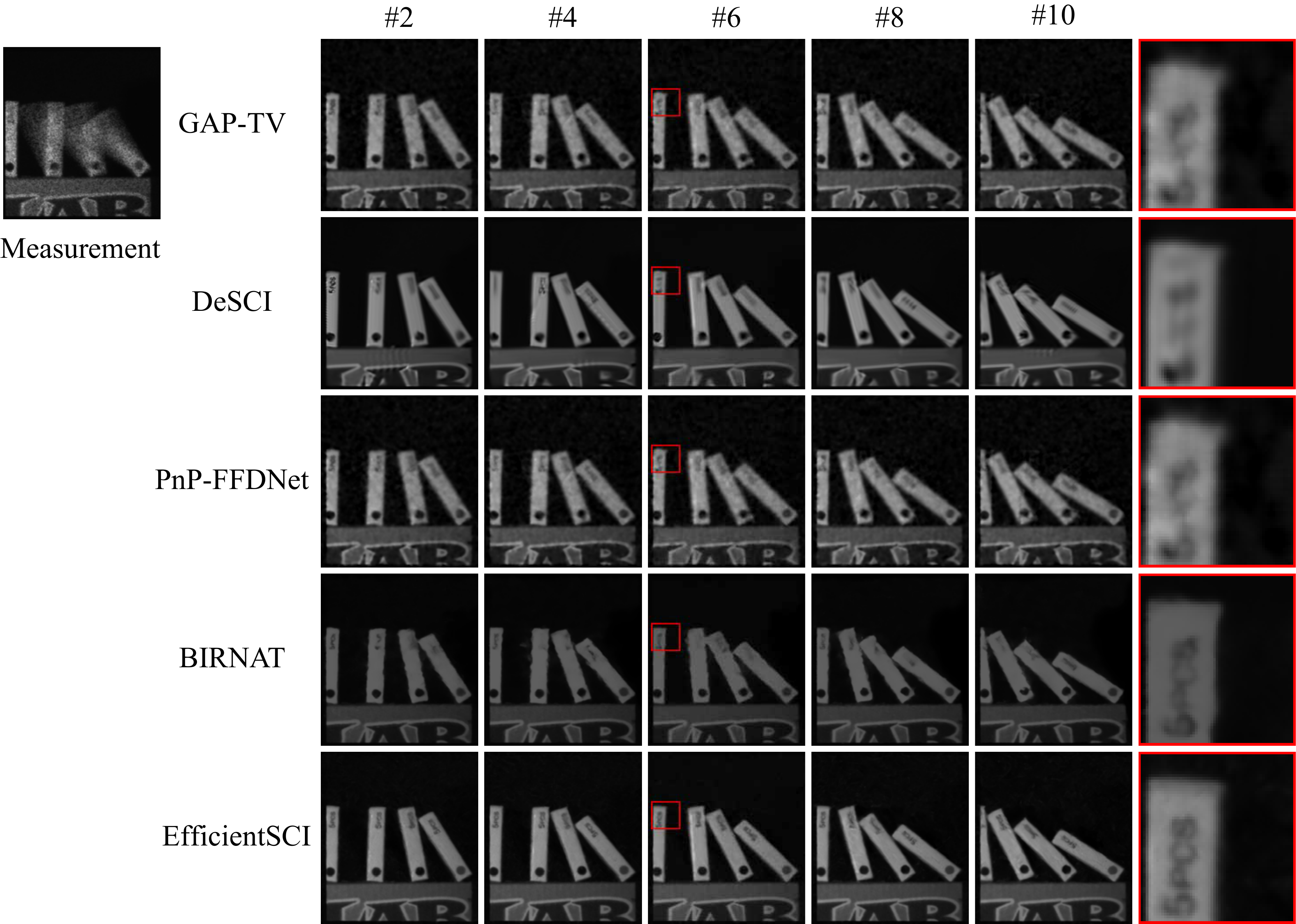}
    \vspace{-5mm}
    \caption{\small{Reconstruction results of different algorithms on \texttt{Domino} real data
      with compression rate $B = 10$.}
    }
  \label{fig:real_cr10}
\end{figure}
\begin{figure}[!ht]
    \centering 
    \includegraphics[width=1.\linewidth]{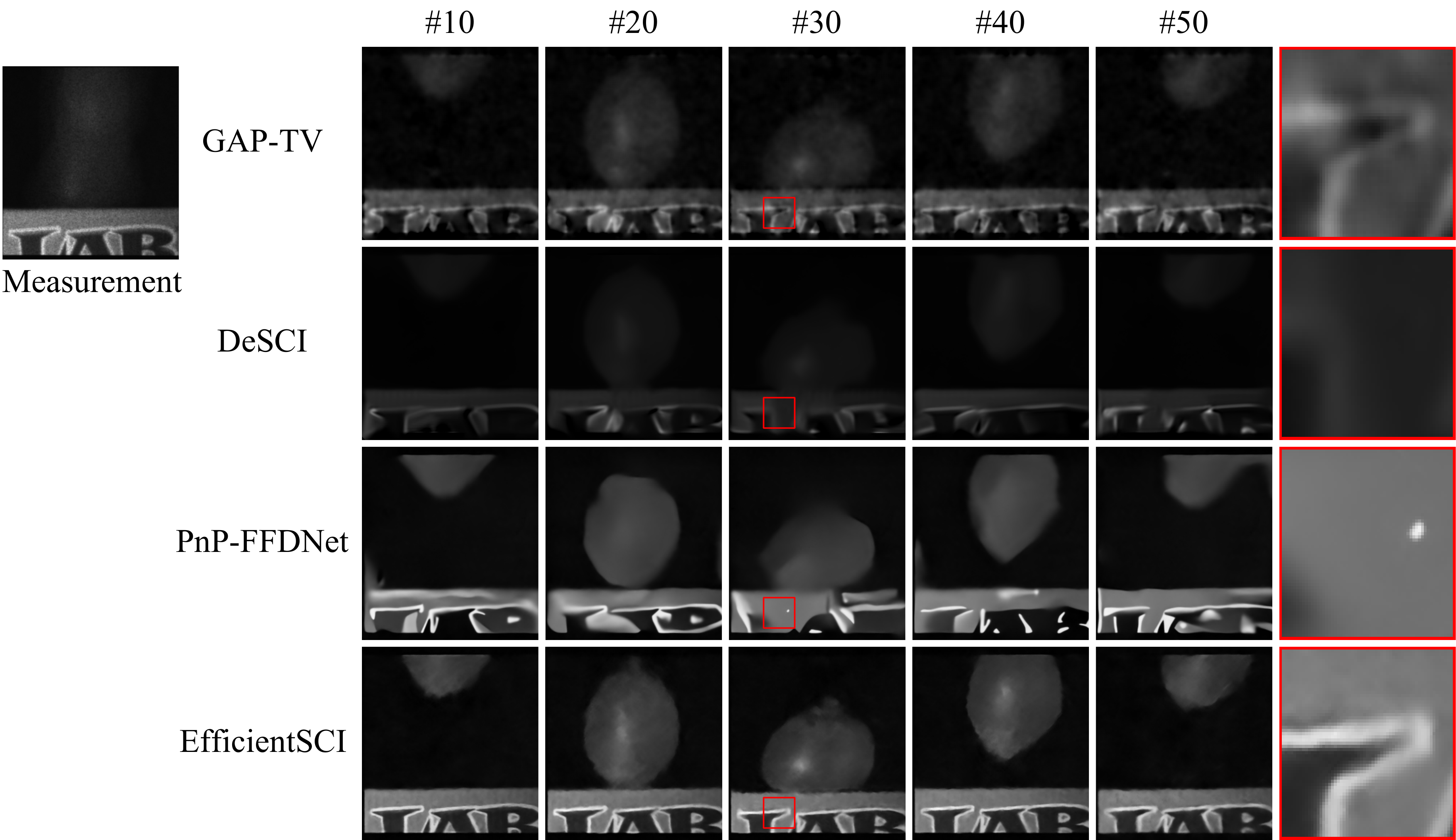}
    \vspace{-5mm}
    \caption{\small{Reconstruction results of different algorithms 
      on \texttt{Water Balloon} real data
      with compression rate $B = 50$.}
    }
    \vspace{-5mm}
  \label{fig:real_cr50}
\end{figure}
\subsection{Results on Real Video SCI Data}

We further test our method on real data. 
Fig. \ref{fig:real_cr10} and Fig. \ref{fig:real_cr50} show the reconstruction results of multiple algorithms on two public data (\texttt{Domino, Water Balloon}), 
we can see that our method can reconstruct clearer details and edges. Specifically, we can clearly recognize the letters on the \texttt{Domino}. 
Even with a high compression ratio ($B=50$), our proposed method can still reconstruct clear foreground and background information (shown in Fig.~\ref{fig:real_cr50}). 

\section{Conclusions and Future Work}
This paper proposes an efficient end-to-end video SCI reconstruction network, dubbed EfficientSCI, 
which achieves the state-of-the-art performance on simulated data and real data, 
significantly surpassing the previous best reconstruction algorithms with high real-time performance. 
In addition, we show for the first time that an UHD color video with high compression rate 
can be reconstructed using a deep learning based method.
For future work, we consider applying EfficientSCI Network to more SCI reconstruction tasks, such as spectral SCI \cite{Yuan2015a,cai2022mask}.

\section*{Acknowledgements}
This work was supported by the National Natural Science Foundation of China [62271414], Zhejiang Provincial Natural Science Foundation of China [LR23F010001]. 
We would like to thank Research Center for Industries of the Future (RCIF) at Westlake University and the funding from Lochn Optics. 
{\small
\bibliographystyle{ieee_fullname}
\bibliography{reference_wangls,reference_xin}
}

\end{document}